\documentclass[lettersize,journal]{IEEEtran}
\usepackage{amsmath,amsfonts}
\usepackage{algorithmic}
\usepackage[bookmarks=true]{hyperref}
\usepackage{graphicx}
\usepackage{epsfig} 
\usepackage{epstopdf} 
\graphicspath{{Figures/}}
\usepackage{array}
\usepackage{textcomp}
\usepackage{stfloats}
\usepackage{url}

\usepackage{verbatim}
\usepackage{amsthm}
\def\BibTeX{{\rm B\kern-.05em{\sc i\kern-.025em b}\kern-.08em
		T\kern-.1667em\lower.7ex\hbox{E}\kern-.125emX}}
\usepackage{balance}
\usepackage{tabularx}
\usepackage[caption=false, font=footnotesize]{subfig}
\usepackage{array}
\usepackage{color}
\usepackage{cite}
\usepackage{lipsum}
\usepackage{makecell}
\usepackage{enumerate}
\newtheorem{theorem}{Theorem}
\newtheorem{remark}{Remark}
\newtheorem{assumption}{Assumption}
\newcommand{\argmin}{\operatornamewithlimits{argmin}}

\begin{document}
\bstctlcite{IEEEexample:BSTcontrol}
	\title{Design and Control of a Bio-inspired Wheeled Bipedal Robot}
	\author{ Haizhou Zhao$ ^{1} $, Lei Yu$ ^{1,2} $, Siying Qin$ ^{1,2} $, Gumin Jin$ ^{3}$, and Yuqing Chen$^{1,*}$
	   \thanks{
		This work was supported by the Jiangsu Science and Technology Program (BK20220283) and the Research Development Fund (RDF-20-01-08). \\
		\indent $ ^{1} $Authors are with Department of Mechatronics and Robotics, School of Advanced Technology, Xi'an Jiaotong-Liverpool University, Suzhou, Jiangsu, China. 
		$ ^{2} $Authors are with University of Liverpool, United Kingdom. 
		$ ^{3} $Authors are with Department of Automation, Shanghai Jiao Tong University, Shanghai, 200240. \\
		\indent This paper has a supplement video material, available at http://ieeexplore.ieee.org.\\
		\indent $^*$E-mail: yuqing.chen@xjtlu.edu.cn
	   }
    }
	\markboth{}%
	{Design and Control of a Bio-inspired Wheeled Bipedal Robot.}
	
	\maketitle

\begin{abstract}
Wheeled bipedal robots (WBRs) have the capability to execute agile and versatile locomotion tasks. This paper focuses on improving the dynamic performance of WBRs through innovations in both hardware and software development. 
Inspired by the human barbell squat, a bionic mechanical design is proposed and implemented as shown in Fig. \ref{fig:overview_first}. It distributes the weight onto its hip and knee joints to improve the effectiveness of joint motors while maintaining a relatively large workspace of the base link. Meanwhile, a novel model-based controller is devised, synthesizing height-variable wheeled linear inverted pendulum (HV-wLIP) model, Control Lyapunov Function (CLF) and whole-body dynamics for theoretically guaranteed stability and efficient computation. Compared with other alternatives, as a more accurate approximation of the WBR dynamics, the HV-wLIP can enable more agile response and provide theory basis for WBR controller design. Experimental results demonstrate that the robot could perform human-like deep squat, and is capable of maintaining tracking CoM velocity while manipulating base states. Furthermore, it exhibited robustness against external disturbances and unknown terrains even in the wild.

\end{abstract}


\section{Introduction}

Wheeled bipedal robots (WBRs) \cite{klemm2019ascento,Ollie} integrates both the advantages of low-cost transportation of wheeled robots and the high trafficability of legged robots. 
However, in recent research, the design and control of WBRs are still primitive compared with the booming development of non-wheeled bipedal robots. In the following sections, we will introduce recent progress on mechanical and controller design of WBRs.

\subsection{Mechanical Design}
The most well-known WBR is the {Ascento} \cite{klemm2019ascento}, developed by ETH  in 2019. The design of {Ascento} includes a pair of four-bar linkage legs, each leg is with 1 degree of freedom (DoF) and equipped with the torsional springs in the knee joint to alleviate impacts. It can jump upstairs or lower down its head to pass obstacles while balancing on wheels. However, its usage is highly limited for complex tasks due to the lack of sufficient DoFs in legs. 

\begin{figure}
	\centering
	\includegraphics[width=0.7\linewidth]{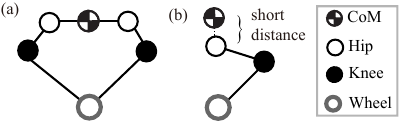}
	\caption{Schematic demonstration of WBRs with 2 DoF legs. (a) Five-bar linkage, (b) planar 2-link serial layout. The dashed line represents the short distance between the CoM and the hip joint, indicating the much longer moment arm of the knee motors.}
	\label{fig:layoutillustrantion}
\end{figure}

Recently, WBRs with 2 DoF legs emerged, adding the ability to control the base height and orientation simultaneously on top of the capability of {Ascento}.  
Similar to robotic manipulators such as delta robots \cite{delta_robotica} or serial robotic arms \cite{irimarm}, the design of WBR legs with 2 DoFs also involves the trade-off between workspace and load capability. A five-bar linkage parallel layout, as is shown in Fig.\ref{fig:layoutillustrantion}a, distributes the weight evenly onto the two symmetrically installed motors to increase load capacity, such as the Tencent {Ollie} \cite{Ollie}. However, the parallel mechanism limits the workspace of the base links, e.g., operational height and base pitch angle. 
\begin{figure}[t]
    \centering
    \includegraphics[width=0.5\linewidth]{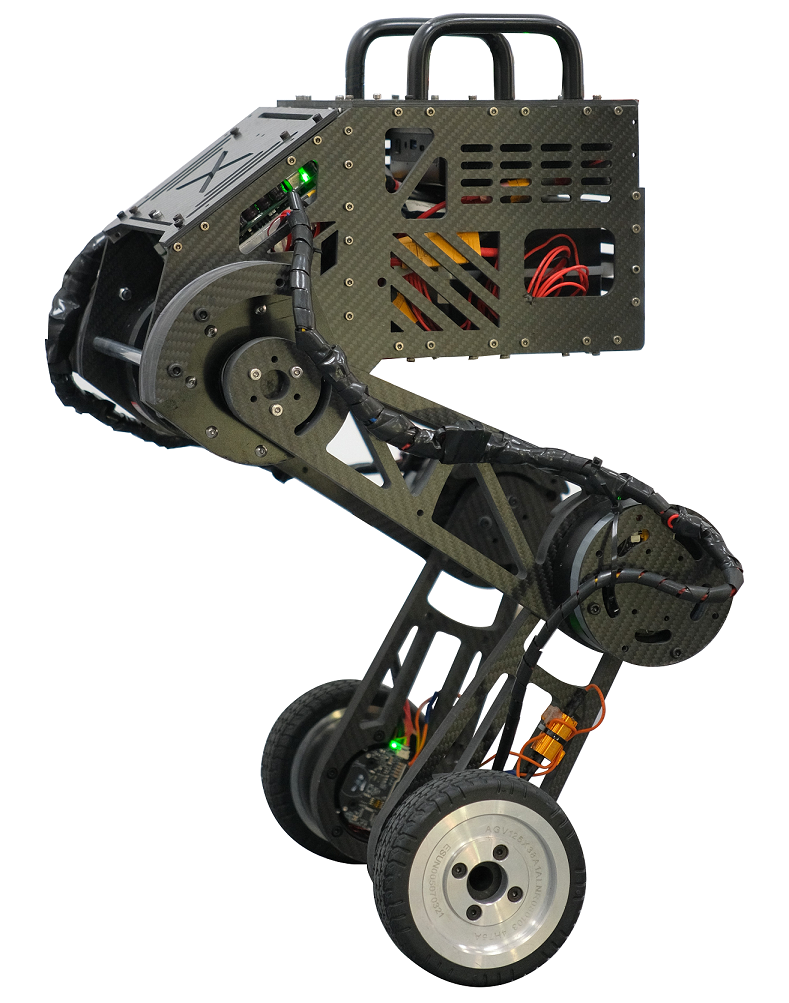}
    \caption{Overview of the bio-inspired wheeled bipedal robot.}
    \label{fig:overview_first}
\end{figure}
On the contrary, WBRs with planar two-link serial legs \cite{sustechnezha, mpc_wl_ral,bitwiptmech,tmechwip}, shown in Fig. \ref{fig:layoutillustrantion}b, have larger workspace. A serial design usually has the knee motor aligned with the hip joint and connected to the knee by a parallel four-bar linkage to reduce the leg inertia. However, this adaptation exhibits low effectiveness in terms of joint torques. For such a WBR in the double support phase, the moment arm of knees are much longer than that of the hips. Consequently, the weight is mainly supported by the knee motors, which usually requires torques higher than the rated value and thus leads to rapid overheat. In contrast, due to the small moment arm, the hip motors contribute much lower torques and can hardly help increase the load capacity.

\subsection{Controller Design}
In previous literature, the controller designed for WBRs are generally based on reduced-order models, of which the modeling precision decides the performance. A widely applied model is the wheeled inverted pendulum (WIP)\cite{RAS_WIP}. It approximates the WBR dynamics by a body-fixed inverted pendulum with wheels. The WIP is integrated with PID controllers \cite{klemm2019ascento, Ollie}, whole body controllers (WBC) \cite{Klemm2020LQRAssistedWC, sustechnezha, tmechwip}, and model predictive controllers (MPC) \cite{mpc_wl_ral, bitwiptmech} for various scenarios and applications. However, to simultaneously control the height and orientation of WBRs with 2-DoF legs, the assumption of fixed body will be inevitably violated. It will not only lead to potential performance degradation, but also invalidate the stability guarantee and model-based controller design due the model mismatch. For example, a WIP-based MPC cannot accurately predict a WBR in acceleration while maintaining a constant base height.

Recent works on WBRs with 2-DoF legs have proposed new models with higher DoFs to overcome the drawbacks of WIP. For instance, a nonlinear wheeled spring-loaded inverted pendulum model (W-SLIP) \cite{sustechnezha} is developed from the bipedal walking theory to improve the performance of nonlinear trajectory optimization, but is inapplicable to real-time control due to its complex nonlinear formulation. A Cart-Linear-Inverted-Pendulum-Model (Cart-LIPM) \cite{Xin2020OnlineDM} is adapted from the linear inverted pendulum (LIP). Its linearity enables real-time trajectory optimization (TO) as MPC and the optimized trajectory is tracked by WBC. However, it does not consider possible center-of-mass (CoM) height variation. Recently, the single rigid body model is adapted to its wheeled version (WRBD) for WBR \cite{mpc_wl_ral}. To manipulate the base velocity and posture, the WRBD is incorporated in nonlinear MPC for online computation of the wheel trajectory and reaction forces which are then tracked by a low-level WBC. The main barrier for application of WRBD is its high nonlinearity, resulting in computational burden by iterative optimization. 
In bipedal walking community, stable controllers based on ankle torques have been widely studied \cite{troheightvariable, angularmomentumcassie}. Despite the similarity between wheel and ankle torques, its fixed contact model cannot describe the rolling contact of wheeled legged robots. 

Besides the model-based control methods, there are works on reinforcement learning (RL) for WBRs control\cite{wheelrl, OLLIE_RL}. These RL methods often require random exploration to collect sufficient training data for learning a control policy, but due to the instability essence of WBRs, random exploration with hardware system is often unavailable. Consequently, RL controllers can only be learned from imprecise simulation, leading to difficulty in transferring the policy to hardware \cite{Recht2019}.

\subsection{Contributions}
In this paper, we propose a bio-inspired WBR with an efficient online controller. The bio-inspired mechanical design could improve the performance of load-carrying squat, which shares similarities to a wide range of challenging tasks in robots with bipedal structures \cite{Xu2018}. Inspired by human deep squats, its bionic design captures the underlying principle of efficient squats and imitates the human mass layout. 
The proposed controller integrates a novel height-variable LIP (HV-wLIP) model with improved model precision. The new model extends the Cart-LIPM \cite{Xin2020OnlineDM} by incorporating the wheel torques and height dynamics; meanwhile, the proposed control method includes a balancing controller based on Control Lyapunov Function (CLF) \cite{Reher2019AnID,khalil2002nonlinear}, to ensure the asymptotic stability of the centroidal self-balancing dynamics.  Advantages of the proposed mechanical design and controllers are demonstrated with comprehensive simulations and hardware experiments. 

The primary contributions could be summarized as: 
\subsubsection{Mechanical Design} We propose a novel bio-inspired WBR design, which overcomes the low torque effectiveness flaw of previous works on serial-legged WBRs while preserving a large workspace. 
\subsubsection{Controller} We propose a novel model-based controller based on the WBC structure. The proposed controller (i) includes the HV-wLIP that incorporates wheel torques and variable height, to improve the control performance and provide theory basis for stability analysis and controller design; (ii) integrates WBC with the CLF-based stability condition and the HV-wLIP model to enable velocity tracking and posture manipulation, and guarantee real-time stability. 
The paper is structured as follows: In Section \ref{sec:bionicdesign}, we introduce the bio-inspired WBR design and the full model dynamics and kinematics. In Section \ref{sec:wlipmodel}, we introduce the HV-wLIP model and propose the CLF stability condition. In Section \ref{sec:controller}, we propose an efficient weighted-QP-based WBC for the WBR. 
In Section \ref{sec:experiment}, we demonstrate the advantages of our model, verify the static and dynamic performance, and the robustness against disturbance of the proposed controller. 
In Section \ref{sec:conclu}, we conclude this paper.

\section{Bio-inspired Mechanical Design}
\label{sec:bionicdesign}
\begin{figure}[t]
	\centering
	\includegraphics[width=0.8\linewidth]{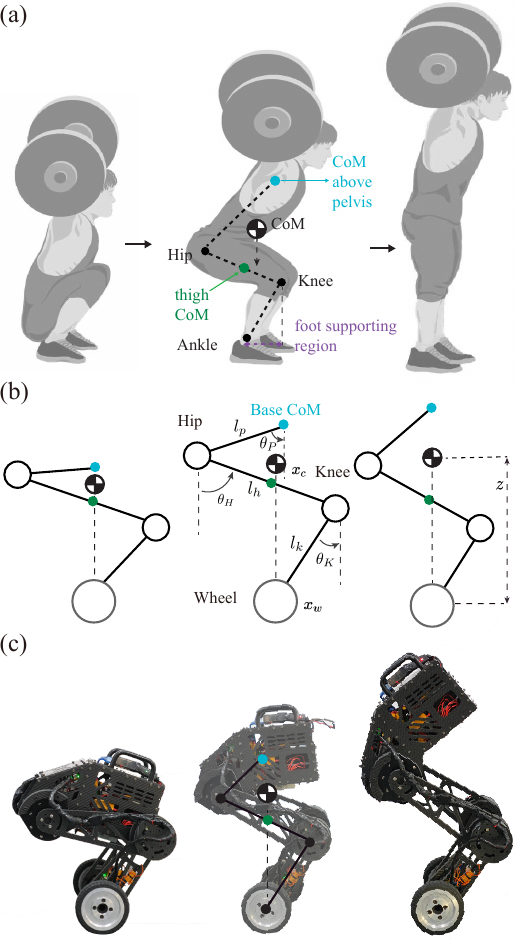}
	\caption{Bionic design of the WBR. (a) Barbell squat illustration of human. (b) The simplified design model imitating human squatting.  (c) The hardware implementation of the bionic WBR during human-like squat. $z$ is the body CoM height from the wheel center and $x_c$, $x_w$ are respectively the body CoM and wheel horizontal positions in sagittal plane. $l_p$, $l_h$, $l_k$ are respectively the link length of the base, thigh, and shank. $\boldsymbol{\Theta}=[\theta_P,\theta_H,\theta_K]$, $\theta_P$, $\theta_H$, and $\theta_K$ are respectively base pitch, hip and knee absolute angles.}
	\label{overview}
\end{figure}

In this paper, we aim to design a bio-inspired WBR with improved squat performance.  
The human bio-mechanics naturally provides a well-generalizing solution to the leg design problem. Therefore, we desire to obtain a suitable WBR design by imitating critical human bio-mechanics. Specially, we note that the squat serves as one of the most common motions in athletic and daily activities, such as load lifting \cite{Javier2020}. To improve the torque effectiveness and increase load capability, we designed our robot by analyzing the typical barbell squat of human demonstrated in Fig. \ref{overview}a.

\subsection{Link Length Design}\label{sec:legdesign}
The link length ratios of our proposed WBR robot are designed to imitate the human torso-thigh-shank ratio. According to the previous results in human bio-mechanics research on powerlifting \cite{legratio} and human body statistics \cite{humandata}, the shank length $l_k$ is approximately 80\% of the thigh $l_h$, and the upper body hip-to-CoM length $l_p$ (excluding head, assuming quasi-uniform mass distribution) is about 60\% of the thigh $l_h$. Consequently, ratio of the leg length of the wheeled-bipedal robot could be similarly defined as $[l_p:l_h:l_k ] = [3:5:4]$.

Note that due to the fundamental difference between human body and WBRs, this ratio is only a rough approximation and does not directly guarantee advantages. By imitating it, we expect the existence of a considerably large feasible set of joint configurations whereby the robot can behave similarly to human squat for potential torque effectiveness improvement. 

We tested candidate link lengths satisfying the above ratio in simulation to find the one providing large enough workspace and ensuring motor torques are normally within the rated range to avoid overheat. Empirically, the final lengths are decided as $l_p=150$mm, $l_h=250$mm, $l_k=200$mm.

\subsection{Mass \& Torque Distribution}\label{sec:bio}
In human bio-mechanics, human leg distributes the weight across hip and knee muscles\cite{Wong2016}. It is also reported that during explosive leg motion such as squatting and jumping, the work done by the hip and knee muscles are approximately the same \cite{Arthur1993, Chen2020BiomechanicalAT}. As illustrated in Fig. \ref{overview}a, when humans attempt to lift a heavy barbell, they will actively change their posture to control the moment arm of each joint to distribute the load onto different muscles. Motivated by these studies, we implemented several features in our design. Firstly, we chose to install properly the auxiliary components, including the batteries, on the base link to ensure $l_p$. The hip and knee motors are placed at the hip and knee joints respectively to imitate the heavy human thigh muscles, as shown in Fig. \ref{overview}b,c. Furthermore, to balance the strength of each joint, we introduce the following torque ratio as a performance index of the torque effectiveness:
\begin{equation} \label{eq:torque_ratio}
	r_\tau=|{\tau_h}/{\tau_k}|,
\end{equation}
where $\tau_h, \tau_k$ are hip and knee joint torques, absolute value aims to eliminate the influence of torque directions. For ease of design, we select the same type of motors with the same rated and maximum output capability. Consequently, during the squat motion, $r_\tau$ should be control around 1 to fully utilize their strength and increase the load capacity. It can be implemented by controlling the moment arm of the load supported by each joint, in the spirit of human deep squats. 
\begin{theorem} (Load Capacity Improvement)
	By controlling $r_\tau=1$, the load capacity of the WBR is maximized for $\mathcal{O}=\{\theta_P,\theta_K\in(-\pi/2, 0), \theta_H \in (0, \pi/2):\theta_H < \theta_P + \pi\}$.
\end{theorem}
\begin{proof}
	Considering the planar 3-link layout in Fig. \ref{overview}(b), in static case the following equation holds for a WBR in balance:
	\begin{subequations}\label{eq:effectivenessproof}
		\begin{align}
			\delta x &= m_cl_k\sin_K+\bar{m_c}l_h\sin_H+m_bl_p\sin_P=0,\\
			z&= m_cl_k\cos_K+\bar{m_c}l_h\cos_H+m_bl_p\cos_P=z^d,\\
			\mathbf{J}_w^\top&\begin{bmatrix}
				0&m_cg
			\end{bmatrix}^\top=\begin{bmatrix}0&\tau_h&\tau_k\end{bmatrix}^\top	,
		\end{align}
	\end{subequations}
	where $\delta x=x_c-x_w$ is the horizontal difference between the CoM and the wheel axis, $z^d$ is the desired CoM height, $\mathbf{J}_w$ is the jacobian that maps $\dot{\boldsymbol{\Theta}}$ to the wheel velocity w.r.t the body CoM in the Cartesian space, $m_c$ is the body mass, $g$ is the gravitational acceleration, $\tau_h, \tau_k$ are respectively the torques of hip and knee motors. 
	
	For the knee joint supporting the body mass, its moment arm is $l_k\sin\theta_K$. For the hip joint supporting the base mass, its moment arm is $l_p\sin\theta_P$. Starting from any $\boldsymbol{\Theta} \in \mathcal{O}$, fixing the base CoM height and changing the pitch angle will change the moment arm of joints. If $|\theta_P|$ is increased, $|l_p\sin\theta_P|$ increases, i.e., the hip torque increases. If $|\theta_P|$ is decreased, i.e., $\delta \theta_P>0$, writing down the first variation of $\delta x$ and $z$:
		\begin{equation}\hspace{-0.15em}
				\begin{bmatrix}
					m_cl_k\cos_K & \bar{m_{c}}l_h\cos_H \\
					m_cl_k\sin_K & \bar{m_{c}}l_h\sin_H 
				\end{bmatrix}\begin{bmatrix}
					\delta \theta_K \\\delta \theta_H
				\end{bmatrix}=-m_bl_p\begin{bmatrix}
					\cos_P\\\sin_P
				\end{bmatrix}\delta \theta_P,
		\end{equation}
		where $\sin_{(\cdot)}=\sin\theta_{(\cdot)},\cos_{(\cdot)}=\cos\theta_{(\cdot)}$, $m_b,m_t$ are respectively the base and thigh mass and $\bar{m_{c}}=m_b + m_t/2$. \newline
			\textbf{Case 1:} For $\theta_H \le \pi/2$, by range of $\Theta$ and $\delta\theta_P>0$, one can derive $\delta \theta_K < 0$, indicating $\theta_K$ is decreased and $|l_k\sin\theta_K|$ is increased, i,e., the knee torque will increase.\newline
			\textbf{Case 2:} For $\theta_H \in (\pi/2, \theta_P+\pi)$, since $\cos_P>\cos[\theta_ H-\pi]=-\cos_H,\sin_P>\sin[\theta_H-\pi]=-\sin_H,\cos_H<0,\sin_H>0$, we have the following inequality system transformed from (4)
			\begin{equation}	
					\begin{bmatrix}
						m_cl_k\cos_K & \bar{m_c}l_h\cos_H \\
						m_cl_k\sin_K & \bar{m_c}l_h\sin_H 
					\end{bmatrix}\begin{bmatrix}
						\delta \theta_K \\\delta \theta_H
					\end{bmatrix}<m_bl_p\begin{bmatrix}
						\cos_H\\\sin_H
					\end{bmatrix}\delta \theta_P.
			\end{equation}
			Dividing $[\cos_H,\sin_H]^\top$ row-wise leads to
			\begin{subequations}
				\begin{align}
					&m_cl_k\cos_K/\cos_H \delta\theta_K + \bar{m_c}l_h \delta \theta_H >m_bl_p\delta\Theta_P,\\
					&m_cl_k\sin_K/\sin_K \delta\theta_K + \bar{m_c}l_h \delta \theta_H <m_bl_p\delta\Theta_P,\vspace{-10em}
				\end{align}
				merging which results in \begin{align}
				\cos_K/\cos_H \delta\theta_K > \sin_K/\sin_H \delta\theta_K\label{eq:case2ineq}.	
				\end{align}
			\end{subequations}
			Since $\cos_K>0,\sin_K>0,$ to make \eqref{eq:case2ineq} feasible, $\delta \theta_K < 0$ is necessary. Therefore, the knee torque increases as in Case 1.
			\par
	 To conclude, for $\boldsymbol{\Theta}\in\mathcal{O}$ and constant $z$, there is a unique solution for $r_\tau=1$ and
	\begin{equation}
		\inf_{\boldsymbol{\Theta}\in\mathcal{O}} {\max}\{|\tau_h|, |\tau_k|\} = \tau_0|_z.
	\end{equation}
	where $\tau_0|_z=|\tau_h|$ when $r_\tau=1$. Assuming the robot weight is scaled by $a>1$, i.e., the $m_c$ and $m_t$ are increased proportionally. $\boldsymbol{\Theta}$ and  $\mathbf{J}_w$ remain the same because the mass layout is not changed. Then \eqref{eq:effectivenessproof} still holds except that the $m_c\to am_c$. By linearity, the right side becomes $\begin{bmatrix}
		0,a\tau_h,a\tau_k
	\end{bmatrix}^\top$. Since the maximum feasible  $a=\tau_{\max}/{\max}\{|\tau_h|, |\tau_k|\}$, the maximal load capacity for a given $z$ is
	\begin{equation}
		{\max}~{am_c}\big|_z=m_{c}\frac{\tau_{\max}}{\inf_{\theta\in\mathcal{O}}{\max}\{|\tau_h|, |\tau_k|\}}=m_{c}\frac{\tau_{\max}}{\tau_0|_z}.
	\end{equation}
	This concludes the proof.
\end{proof}
To the best of our knowledge, there have been many successfully implemented wheeled bipedal robots while none has explicitly considered this from the aspect of bionic inspiration. 
\subsection{Dynamic Model}
\label{sec:modeldynamics}

\begin{figure}[t!]
	\centering
	\includegraphics{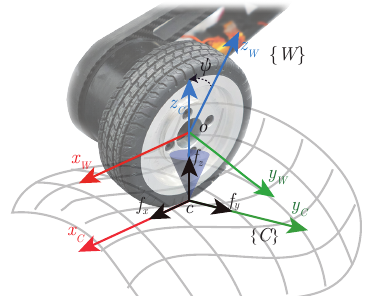}
	\caption{Contact frame illustration of the right wheel. $\{C\}$ has its z-axis normal to the ground, its x-axis opposite the floating-base x-axis, and its origin at the contact point $c$. $\{W\}$ is rotated from $\{C\}$ by clockwise angle $\psi$ w.r.t. $x_C$ and has its origin at the wheel center $o$. The transparent blue cone represents the friction cone constraint of the 3D linear force components $\mathbf{f}_c^{(\cdot)}=[f_x,f_y,f_z]^\top$ where $(\cdot)$ can be left or right.}
	\label{fig:contact}
\end{figure}

The proposed WBR shown in Fig. \ref{overview}c has 6 actuated DoFs (joints and wheels for two legs) $\mathbf{q}_j\in\mathbb{R}^6$ and 6 floating DoFs, i.e., XYZ translation $\mathbf{r}_\text{xyz}\in\mathbb{R}^3$ and Euler ZYX rotation: yaw $\phi\in\mathbb{R}$, pitch $\theta\in\mathbb{R}$, roll $\psi\in\mathbb{R}$.   Note that in this project, we use Featherstone's spatial algebra notation \cite{featherstone2014rigid}. Let $\mathbf{q}=[\mathbf{r}_{xyz}, \phi, \theta, \psi, \mathbf{q}_j]^\top\in\mathbb{R}^{12}$, the robot dynamics can be expressed in the following equation of motion (EoM)
\begin{subequations}\label{eq:EoM}
\begin{align}
\mathbf{M}(\mathbf{q})\ddot{\mathbf{q}}+\mathbf{H}(\mathbf{q},\dot{\mathbf{q}})&=\mathbf{S}\boldsymbol{\tau}+{}_C\mathbf{J}^\top_c{}_C\mathbf{F}_c,\\
{}_W\mathbf{J}_c\ddot{\mathbf{q}}+{}_W\dot{\mathbf{J}}_c\dot{\mathbf{q}}&=\mathbf{C}_r,
\label{eq:contactconstr}
\end{align}
\end{subequations}
where $\mathbf{M}\in\mathbb{R}^{12\times12}$ is the joint space inertia matrix (JSIM), $\mathbf{H}\in\mathbb{R}^{12}$ is the nonlinear effect including the Coriolis force and gravitational force, $\mathbf{S}=[\mathbf{0}_{6\times6},\mathbf{I}_{6}]^\top\in\mathbb{R}^{12\times6}$ is the selection matrix, $\boldsymbol{\tau}\in\mathbb{R}^6$ is the driven torque in each motor, ${}_W\mathbf{J}_c$, ${}_C\mathbf{J}_c\in\mathbb{R}^{6\times12}$ are respectively the contact Jacobian matrices expressed in $\{W\}$ and $\{C\}$ of the two wheels as illustrated in Fig. \ref{fig:contact}, ${}_C\mathbf{F}_c=[\mathbf{f}_c^\text{left}, \mathbf{f}_c^\text{right}]\in\mathbb{R}^{6}$ is the vector of the linear contact forces exerted on $c$, and $\mathbf{C}_r\in\mathbb{R}^{6}$ is the rolling constraint defined later. 
\subsubsection*{Rolling Constraints}
To estimate the base linear velocity $\dot{\mathbf{r}}_{xyz}$ and enable control based on inverse dynamics (ID), we need to utilize the contact constraints.
Let $\boldsymbol{\omega}_o\in\mathbb{R}^3$ be the wheel angular velocity expressed in $\{W\}$. The rolling kinematic constraint refers to maintaining the linear velocity of the contact point $c$ as zero, i.e.,
\begin{equation}\label{eq:kincon}{}_W\mathbf{v}_c=
\begin{bmatrix}
\boldsymbol{\omega}_o\\
\mathbf{0}_{3\times1}
\end{bmatrix}=\begin{bmatrix}
\mathbf{I}_3&\mathbf{0}_{3\times3}\\
-{}_W\mathbf{r}_c\times&\mathbf{I}_3
\end{bmatrix}\begin{bmatrix}
    \boldsymbol{\omega}_o\\
    {}_W\boldsymbol{v}_{o}
\end{bmatrix},
\end{equation}
where $o$ is the wheel center, ${}_W\textbf{r}_c=[0, 0, -r_w]$ is the contact point position and $r_w$ is the wheel radius. By differentiating the linear velocity part in \eqref{eq:kincon} with respect to time, we have
\begin{equation}\label{eq:zeroacc}
    \mathbf{0}_{3\times1}=-{}_W\mathbf{r}_c\times\dot{\boldsymbol{\omega}}_o+{}_W\dot{\boldsymbol{v}}_{o}.
\end{equation}
Let $\sigma$ be the counterclockwise angle between the x-axis of $\{W\}$ and $p_\sigma$ be the position of points on the wheel contour parameterized by $\sigma$. The linear spatial acceleration of $p_\sigma$ expressed in $\{W\}$ is
\begin{equation}\label{eq:spatialacc}
	{}_W\dot{\boldsymbol{v}}_{p_\sigma} = -\dot{\mathbf{r}}(\sigma) \times \boldsymbol{\omega}_o-\mathbf{r}(\sigma) \times \dot{\boldsymbol{\omega}}_o+\dot{{}_W\boldsymbol{v}}_{o},
\end{equation}
where $\mathbf{r}(\sigma)=[r_w\cos\sigma, 0, -r_w\sin\sigma]^\top\in\mathbb{R}^3$ is the displacement from the wheel center to the point. For the contact point, $c=p_{\pi/2}$, $\sigma=\pi/2, \mathbf{r}(\sigma)={}_{W}\textbf{r}_c$. As $\dot{\mathbf{r}}(\sigma)=[-r_w\dot{\sigma}\sin\sigma,0,-r_w\dot{\sigma}\cos\sigma]^\top$, substituting \eqref{eq:zeroacc} into \eqref{eq:spatialacc} gives
\begin{equation}\hspace{-0.5em}
{}_{W} \dot{\boldsymbol{v}}_c=-\dot{\mathbf{r}}(\pi/2) \times \boldsymbol{\omega}_o =[
0,-r_w \dot{\sigma}\omega_{o z},r_w \dot{\sigma} \omega_{o y}	
]^\top\in\mathbb{R}^3,
\end{equation}
where $\dot{\sigma}$ is the time derivative of the wheel pitch angle, and the $\omega_{oz}, \omega_{oy}$ are the $z, y$ components of $\boldsymbol{\omega}_o$, which can be calculated using \textit{Pinocchio} \cite{Carpentier2019Pinocchio}. Consequently, the rolling constraint $\mathbf{C}_r$ in \eqref{eq:contactconstr} is defined as
$\mathbf{C}_r=[{}_{W} \dot{\boldsymbol{v}}_{c,\text{left}}^\top,{}_{W}\dot{\boldsymbol{v}}_{c,\text{right}}^\top]^\top.$
\subsubsection*{Linearized Friction Cone}
To avoid slippage of wheels and consequent failure in contact-aided state estimation, friction cone constraint is often added, i.e.,
\begin{equation}\label{eq:orignalfriccon}
	\forall \mathbf{f}_c^{(\cdot)} \in {}_C\mathbf{F}_c,\;f_z^{(\cdot)}\ge 0, \;
		\sqrt{[f_x^{(\cdot)}]^2+[f_y^{(\cdot)}]^2}
	 < \mu f_z^{(\cdot)},
\end{equation}
where ${}^{(\cdot)}$ can be left or right, and $\mu$ is the friction coefficient.
In common WBC \cite{Bellicoso2016HQP,Klemm2020LQRAssistedWC}, \eqref{eq:orignalfriccon} is linearized for QP framework:
\begin{equation} \label{eq:friccon}
	\forall \mathbf{f}_c^{(\cdot)} \in {}_C\mathbf{F}_c,\;f_z^{(\cdot)}\ge 0, \;|f_x^{(\cdot)}| < \mu f_z^{(\cdot)},\; |f_y^{(\cdot)}| < \mu f_z^{(\cdot)}.
\end{equation}

\section{Reduced-Order Model}
\label{sec:wlipmodel}
\begin{table}[t]
	\centering
	\caption{Model Assumptions}
	\setlength{\tabcolsep}{2.5mm}{\begin{tabular}{lllll}
			\hline\hline\noalign{\smallskip}	
			\textbf{Name} & \textbf{Leg} & \textbf{Body} & \textbf{Contact} & $m_w=0?$ \\
			\noalign{\smallskip}\hline\hline\noalign{\smallskip}
			WIP & fixed $l$ & rigid link & rolling&no\\
			LIP & constant $z$ & zero CAM & fixed&-\\
			HV-wLIP & variable $z$ & zero CAM & rolling&no\\
			\noalign{\smallskip}\hline\hline
	\end{tabular}}  
	\label{tab:modelassumption}
\end{table}
To control the WBR, we propose a reduced-order model HV-wLIP. For comparison, we introduce the WIP, LIP and HV-wLIP respectively and conclude their model assumptions in Table. \ref{tab:modelassumption}. An infinite-horizon optimal control problem (OCP) is formulated to derive a time-varying CLF. To integrate with the optimization-based control scheme, a CLF-based runtime stability guarantee is devised. 
\subsection{The WIP Model}
We first recall the WIP as the baseline for comparison. The WIP \cite{RAS_WIP} as shown in Fig. \ref{fig:simpmodel}i is widely used for two-wheeled robot. It assumes that the body (all parts excluding the wheels) is a rigid link by fixing all internal DoFs such that it will not change length and inertia. One can derive its EoM by the Euler-Lagrange Equation as
\begin{equation}
	\begin{aligned}
		&\begin{bmatrix}
			I_c+m_cl^2 & m_cl\cos\vartheta\\
			m_clr_w \cos\vartheta & \frac{I_w+(m_c+m_w)r_w^2}{r_w}
		\end{bmatrix} \begin{bmatrix}
			\ddot{\vartheta}\\\ddot{x}_w
		\end{bmatrix}\\&\;\;\;\;\;\;\;\;\;\;\;\;\;\;\;\;\;\;\;\;\;\;\;\;\;\;\;-m_cl\sin\vartheta\begin{bmatrix}
			g\\
			r_w\dot{\vartheta}^2
		\end{bmatrix}
		=\tau_w\begin{bmatrix}
			-1\\1
		\end{bmatrix},
	\end{aligned}
\label{eq:wip}
\end{equation}
where $g$ is the gravity acceleration, $l$ is the distance between the CoM and the wheel center, $r_w$ is the wheel radius, $I_w$ is the wheel centroidal inertia, $I_c$ is the body centroidal inertia, $\tau_w$ is the wheel torque, $x_w$ is the wheel center position, and $\vartheta$ is the clockwise CoM angle between the gravity and the body-wheel linkage. $m_c$ is the body mass and $m_w$ is the wheel mass. 

\subsection{The HV-wLIP Model}\label{sec:wlip}
\begin{figure}[t]
	\centering
	\includegraphics[width=0.8\linewidth]{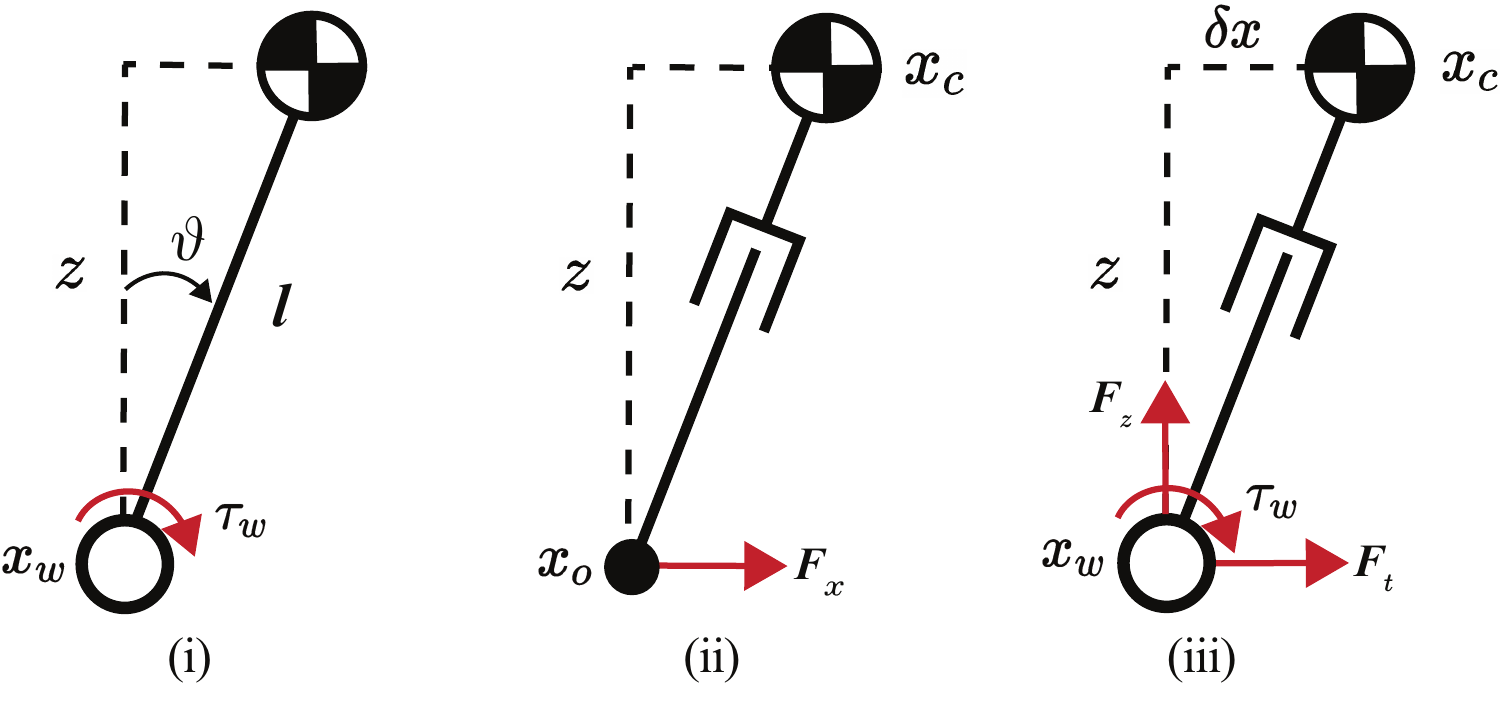}
	\caption{Simplified model illustration. (i) WIP (ii) LIP (iii) HV-wLIP.}
	\label{fig:simpmodel}
\end{figure}

We first recap the Linear Inverted Pendulum (LIP) model as shown in Fig. \ref{fig:simpmodel}ii, it is a well-studied reduced-order model of bipedal robot walking \cite{kajitalip}. The LIP model constrains the CoM height to be constant and that the centroidal torque should be zero to ensure zero centroidal angular momentum (CAM):
\begin{equation}
F_x=m_cg\frac{x_c-x_0}{z},
\end{equation}
where $F_x$ is the horizontal component of the ground reaction force (GRF), $m_c$ is the body mass, $z$ is the CoM height, $x_c$ is the CoM horizontal position, and $x_0$ is the stance foot location.
Therefore, the CoM planar dynamics is
\begin{equation}
    \ddot{x}_c=\frac{g}{z}(x_c-x_0).
\end{equation}
The LIP, named after its linearity, significantly facilities the controller design and stability analysis of walking robots. However, it is not applicable to WBRs because of the fundamental difference between wheeled-legged and purely legged locomotion, i.e., the time-varying contact position.

To adapt the idea of LIP for WBRs, we first recognize that a horizontal reaction force $F_t$ between the body and the wheel replaces the GRF $F_x$. We keep the zero CAM constraints but allow varying CoM height, leading to the HV-wLIP as shown in Fig. \ref{fig:simpmodel}iii with the following EoM:
\begin{subequations}\label{eq:hvwlipeom}
\begin{align}
&\text{CAM constraint}:zF_t+\tau_w=F_z(x_c-x_w),\\
&\text{Newton's 2nd Law}:\frac{\tau_w}{r_w}-F_t=m_w\ddot{x}_w,F_t=m_c\ddot{x}_c
\end{align}
\end{subequations}
where $F_z=m_c(g+\ddot{z})$ is the vertical reaction force between the wheels and legs.
Let $\delta x=x_c-x_w$, \eqref{eq:hvwlipeom} can be rewritten as the following second-order system:
\begin{equation}\label{eq:wlip}
\begin{aligned}   
\begin{bmatrix}
\ddot{x}_c\\\delta\ddot{x}\\\delta\dot{x}
\end{bmatrix}=&\underbrace{\begin{bmatrix}
0&0&\gamma\\
0&0&\alpha\gamma\\
0&1&0
\end{bmatrix}}_{\mathbf{A}(t)}
\begin{bmatrix}
\dot{x}_c\\\delta\dot{x}\\\delta{x}
\end{bmatrix}+\underbrace{\begin{bmatrix}
-\zeta\\-\alpha\zeta-\beta\\0
\end{bmatrix}}_{\mathbf{B}(t)}\tau_w,
\end{aligned}
\end{equation}
where $\gamma= [g+\ddot{z}(t)]/z(t)$, $\alpha=1+m_c/m_w$, $\zeta= 1/[m_cz(t)]$, and $\beta=1/(m_wr_w)$.

\begin{remark}[Model Accuracy of HV-wLIP] \label{rem:wlip_accuracy}

	Due to its nonlinearity, WIP is commonly linearized around $\vartheta=0$ for linear controllers such as LQR \cite{klemm2019ascento}. For height manipulation, normally parameters (inertia, legs) of the reference model is modified by forward kinematics online \cite{Klemm2020LQRAssistedWC,bitwiptmech}, which conflicts with the fixed-body assumption. The WIP-based stability analysis does not hold for such modification and tuning of controllers lacks theoretical basis. On the contrary, the HV-wLIP models full centroidal dynamics of a WBR with variable height and is insensitive to inertia variation. It can be seen as a time-varying linear system, providing adequate model fidelity and convenience in controller design. 
\end{remark}
\begin{remark}[Dynamic Performance of HV-wLIP]\label{rem:wlip_perf}
HV-wLIP enables better dynamic performance during sliding with the CAM constraint. A similar idea has been introduced by biomechanics study \cite{herr2008angular} and applied in bipedal walking control \cite{gong2021angular, highspeed}. While the WIP requires whole body to rotate during velocity regulation, the HV-wLIP decouples upper body from movements of the legs, leading to lower CAM. With less excessive kinetic energy to damp out, the HV-wLIP enables more highly dynamic tasks such as agile velocity control. 
\end{remark}
\begin{remark}[Extensibility of HV-wLIP]\label{rem:wlip_exten}
The HV-wLIP model balances by modulating $\delta x$, as is shown in Fig. \ref{fig:simpmodel}iii, rendering it suitable for kinematic constraints such as step length bound using tools like Control Barrier Functions (CBFs) in \cite{ames2019cbf}.	
\end{remark}

\subsection{HV-wLIP Infinite-Horizon OCP}
The nonlinear system \eqref{eq:wlip} can be transformed into time-varying linear system with a planned trajectory of $z(t)$ including its time-derivatives. A infinite-horizon OCP can then be formed for the time-varying linear system to track constant CoM velocity reference $\dot{x}_c^d$:
\begin{subequations}\label{eq:ocp}
	\begin{align}
		\min_{\tau_w(t)}~&\frac12\int_0^\infty\mathbf{e}(t)^\top\mathbf{Q}\mathbf{e}(t)+R\tau_w(t)^2dt\\
		\text{s.t.}~&\mathbf{e}=\begin{bmatrix}
			\dot{x}_c-\dot{x}_c^d&\delta\dot{x}&\delta{x}
		\end{bmatrix}^\top,\label{eq:state}\\
		&\dot{\mathbf{e}}=\mathbf{A}(t)\mathbf{e}+\mathbf{B}(t)\tau_w,\label{eq:clsys}
	\end{align}
\end{subequations}
where $\mathbf{Q}$ is a positive definite state running cost weight matrix, $R>0$ is the scalar input cost weight. Inspired by the handling of variable height in \cite{troheightvariable}, we made an assumption about the planned height trajectory:
\begin{assumption}
	$\exists t_s>0$, $\forall t\ge t_s$, $z(t)=z^d$ and $\ddot{z}=0$.
\end{assumption}
With this assumption, the OCP in \eqref{eq:ocp} can be transformed into a combination of a finite-horizon time-varying LQR (TVLQR) problem parameterized by $z(t),\ddot{z}(t),0<t<t_s$ and an infinite-horizon LQR problem with $z=z^d,\ddot{z}=0$ for $t>t_s$. Then \eqref{eq:ocp} can be transformed into the following finite-horizon linear predictive controller:
\begin{equation}\label{eq:TVLQR}
	\hspace{-0.65em}
	\begin{aligned}\min_{\tau_w(t)}~&\mathbf{e}(t_s)^\top\bar{\mathbf{P}} \mathbf{e}(t_s)+\frac{1}{2}\int_{0}^{t_s}[\mathbf{e}(t)^\top\mathbf{Q}\mathbf{e}(t)+R\tau_w(t)^2]dt,\\
		\text{s.t.}~&\eqref{eq:state}-\eqref{eq:clsys}.
	\end{aligned}
\end{equation}
where $\bar{\mathbf{P}}$ is the matrix in the quadratic value function $V(\mathbf{e})=\mathbf{e}^\top\bar{\mathbf{P}}\mathbf{e}$ calculated from the continuous-time algebraic Riccati equation (CARE)
\begin{equation}\label{eq:infriccati}
	(\mathbf{A}^\top\tilde{\mathbf{P}}+\tilde{\mathbf{P}}\mathbf{A}-\tilde{\mathbf{P}}\mathbf{B}R^{-1}\mathbf{B}^\top\tilde{\mathbf{P}})\big|_{t_s}+\mathbf{Q}=\mathbf{0},
\end{equation}
for the $z^d$-parameterized infinite-horizon LQR. The predictive controller \eqref{eq:TVLQR} can be solved by efficient backward integration of the following continuous-time Riccati differential equation
\begin{equation}\label{eq:finitericcati}
	(\mathbf{A}^\top\mathbf{P}+\mathbf{P}\mathbf{A}-\mathbf{P}\mathbf{B}R^{-1}\mathbf{B}^\top\mathbf{P})\big|_t+\mathbf{Q}=-\dot{\mathbf{P}}(t),
\end{equation}
with the boundary condition $\mathbf{P}(t_s)=\tilde{\mathbf{P}}$, where the value function is in the form of $V(t,\mathbf{e})=\mathbf{e}^\top\mathbf{P}(t)\mathbf{e}$. For $z$ close to $z^d$, the TVLQR can be dropped and only the infinite-horizon LQR is needed.
\begin{remark}
	(Solver Implementation) Due to its linear nature, the TVLQR requires only one iteration to solve. In our C++ implementation, we use the riccati recursion to compute for 100 shooting nodes within 10us. The infinite-horizon LQR, however, is calculated using \cite{care_solver} within 25$\mu s$.
\end{remark}
\subsection{CLF-based Stability Analysis} \label{sec:clf}

It is shown in \cite{khalil2002nonlinear} that for a the time-varying linear system in \eqref{eq:clsys},
we can find a quadratic-form scalar CLF $V(t,\mathbf{x})$ bounded by two class $\mathcal{K}_\infty$ functions $\alpha_\text{lb}(||\mathbf{x}||), \alpha_\text{ub}(||\mathbf{x}||)$, i.e.,
\begin{equation} \label{eq:lya_ub_lb}
\forall t\ge0,~\alpha_\text{lb}(||\mathbf{x}||)\leq V(t,\mathbf{x}) \leq \alpha_\text{ub}(||\mathbf{x}||),
\end{equation}
where $\alpha(\cdot)$: $[0,\infty)\to\mathbb{R}_+$ is a monotonically increasing function.  

Besides, the temporal derivative  $\dot{V}(\mathbf{x},\mathbf{u})$
is bounded as 
\begin{equation}\label{eq:decbound}
\forall t\ge0,~	\inf_{u}\dot{V} (t,\mathbf{x},\mathbf{u})\leq -\alpha_{\inf} (||\mathbf{x}||).
\end{equation}
The CLF provides global exponential stability.
With the bounded property in (\ref{eq:decbound}), CLF can be naturally incorporated into optimization-based control to guarantee stability under various constraints such as actuator limits.

With the derived close-loop error system \eqref{eq:clsys}, the stability condition could be analyzed with following theorem.
\begin{theorem} (CLF for HV-wLIP).
	Controllers satisfying the  inequality condition could guarantee the stability of \eqref{eq:clsys}
	\begin{equation}\label{eq:clf}
		\exists \lambda>0, \;\mathbf{e}^\top(\mathbf{P} + \mathbf{P}^\top)[\mathbf{A}\mathbf{e}+\mathbf{B}\tau_w]\leq-\lambda ||\mathbf{e}||_2^2, 
	\end{equation}
where $\mathbf{P}$ is the positive definite solution of the OCP \eqref{eq:ocp} at the current timestep $t=0$.
	
\end{theorem}
\begin{proof}
	For the linear system in (\ref{eq:clsys}), the quadratic function $V$ for both TVLQR and infinite-horizon LQR can be easily proven to satisfy the condition \eqref{eq:lya_ub_lb}. For timesteps with TVLQR, by \eqref{eq:finitericcati}, one can derive that
	\begin{equation}
		\dot{V}(t,\mathbf{e},\tau_w)=-\mathbf{e}^\top\underbrace{(\mathbf{Q}+\mathbf{P}\mathbf{B}R^{-1}\mathbf{B}^\top\mathbf{P})}_{\mathbf{C}(t)}\mathbf{e}.
	\end{equation} 
	Since $\mathbf{P}\mathbf{B}R^{-1}\mathbf{B}^\top\mathbf{P}$ is positive semi definite and $\mathbf{Q}$ is positive definite, i.e., $		\mathbf{e}^\top\mathbf{Qe}>0,\mathbf{e}^\top\mathbf{P}\mathbf{B}R^{-1}\mathbf{B}^\top\mathbf{P}\mathbf{e}\ge0,$ we have 
	 $\mathbf{C}(t)$ is positive definite. Same routine can be applied to the infinite-horizon LQR value function with $\dot{V(\mathbf{e},\tau_w)}=-\mathbf{e}^\top\tilde{\mathbf{C}}\mathbf{e},\tilde{\mathbf{C}}=\mathbf{Q}+\tilde{\mathbf{P}}\mathbf{B}R^{-1}\mathbf{B}^\top\tilde{\mathbf{P}}|_{t_s}$. Therefore, the upper bound of the Lyapunov Function time derivative at each timestep can be found for every any $t$:
	\begin{equation}
		\label{eq:condition}
		\dot{V}(t,\mathbf{e},\tau_w)=-\mathbf{e}^\top\mathbf{C}(t)\mathbf{e}\le-\lambda_\text{min}(\mathbf{C}(t))||\mathbf{e}||_2^2.
	\end{equation}
	where $\lambda_\text{min}(\cdot)$ denotes the minimum eigenvalue of $\mathbf{C}(t)$. To obtain the global exponential stability, selecting for  (\ref{eq:decbound})
	\begin{equation}\label{eq:lambda}
		\lambda=\inf_t
		\begin{cases}
			\lambda_{\min}\mathbf{C}(t), &0\le t<t_s,\\
			\lambda_{\min}\tilde{\mathbf{C}},&t\ge t_s,
		\end{cases}
	\end{equation}such that $\lambda||\mathbf{e}||_2^2$ is a class $\mathcal{K}_\infty$ function and $\dot{V}\le-\lambda||\mathbf{e}||_2^2,\forall t\ge0$. Checking condition \eqref{eq:lya_ub_lb}-\eqref{eq:decbound}, $V(t,\mathbf{e})$ is a CLF function that guarantees asymptotic stability.	Subsequently, we note that
	\begin{equation}
		\label{eq:transition}
	\dot{V}(t,\mathbf{e},\tau_w)=\frac{\partial V}{\partial \mathbf{e}}\dot{\mathbf{e}}=\mathbf{e}^\top(\mathbf{P} + \mathbf{P}^\top)[\mathbf{A}\mathbf{e}+\mathbf{B}\tau_w].
	\end{equation}
	Substituting \eqref{eq:transition},\eqref{eq:lambda} into \eqref{eq:condition} gives the condition in \eqref{eq:clf}. As all the derivations above are equivalent relation, we note that given condition \eqref{eq:clf}, the CLF stability conditions \eqref{eq:lya_ub_lb}-\eqref{eq:decbound} are satisfied, and then the closed-loop error system \eqref{eq:clsys} is stable. This concludes the proof.	
\end{proof}

\begin{remark} [Relaxed CLF condition]
In engineering practice, to improve numerical stability and computational efficiency, we would introduce the relaxed CLF inequality constraint as 
\begin{equation}\label{eq:relaxedclfineq}
	\mathbf{e}^\top(\mathbf{P} + \mathbf{P}^\top)[\mathbf{A}\mathbf{e}+\mathbf{B}\tau_w]\leq-\lambda ||\mathbf{e}||_2^2 + s,
\end{equation}
where $s\geq0$ is a slack variable which will be minimized numerically.

\end{remark}

\section{Tracking Controller Design} \label{sec:controller}

In this section, we design a real-time QP controller that tracks the desired CoM height $z^d$ and base orientation $\theta^d$, $\phi^d$, $\psi^d$. With various constraints to be added to the controller, including contact constraints, actuator constraints and full-order dynamics, we incorporate the relaxed stability condition \eqref{eq:relaxedclfineq} to track the desired CoM velocity $\dot{x}_c^d$.

\subsection{Model Consistency}\label{sec:wipconsis}
To ensure HV-wLIP is a good approximation to the real system, it is important to impose virtual constraints on the CAM. In our case, the sagittal CAM is composed of the leg and base momentum. Due to the limited DoF per leg, the direct control of CAM conflicts with the base pitch manipulation. Instead, we control solely the pitch by a PD law, since its inertia accounts for the majority.

We apply the following PD controller for the CoM height
\begin{equation}
	\label{eq:comz_task}
    \mathbf{J}_z\ddot{\mathbf{q}}=\mathbf{K}_{z}\begin{bmatrix}{z}^d-{z}\\\dot{z}^d-\dot{z} \end{bmatrix}-\dot{\mathbf{J}}_z\dot{\mathbf{q}},
\end{equation}
where $\mathbf{J}_{z}$ is the CoM height Jacobian, $\dot{\mathbf{J}}_{z}\dot{\mathbf{q}}$ is the bias, $\mathbf{K}_{z}$ is the task PD gain matrix and ${z}^d,\dot{z}^d$ are respectively the CoM height reference and its velocity.

The aforementioned base pitch task is formulated as 
\begin{equation}
	\label{pel_pit}
	J_{\theta} \ddot{\mathbf{q}}=K_\theta(\theta^d-\theta)+K_{\dot{\theta}}(-\dot{\theta}),
\end{equation}
where $J_\theta$ is the pitch Jacobian which is a constant matrix, $\theta^d$ is the user commanded pitch reference, $K_\theta$, $K_{\dot{\theta}}$ are respectively the position and velocity gains. Similarly, the yaw ($\phi$) and roll ($\psi$) tracking task is formulated as
\begin{equation}
	\begin{bmatrix}
		J_{\phi}\\
		J_{\psi} 
	\end{bmatrix} \ddot{\mathbf{q}}=\begin{bmatrix}
		K_\phi(\phi^d-\phi)+K_{\dot{\phi}}(-\dot{\phi})\\
		K_\psi(\psi^d-\psi)+K_{\dot{\psi}}(-\dot{\psi})
	\end{bmatrix}.
	\label{yaw_roll}
\end{equation}

During movement, the HV-wLIP requires the legs aligned with each other, while the two legs tend to depart due to the ground reaction force tangential to the moving direction. Although the model would still work for moderate departure when the whole CoM is regulated around the narrow support region between wheels. In our WBC, it is conservatively restricted to zero by a task space PD control law as
\begin{equation}\label{eq:sync}
    (\mathbf{J}_{w,r}-\mathbf{J}_{w,l})\ddot{\mathbf{q}}=\mathbf{K}_{w}\begin{bmatrix}{x}_{w,r}-x_{w,l}\\\dot{x}_{w,r}-\dot{x}_{w,l}\end{bmatrix} - (\dot{\mathbf{J}}_{w,r} -\dot{\mathbf{J}}_{w,l})\dot{\mathbf{q}},
\end{equation}
where $\mathbf{J}_{w,(\cdot)}$ is the right/left wheel Jacobian, $x_{w,(\cdot)}, \dot{x}_{w,(\cdot)}$ is the horizontal right/left wheel position projected onto the sagittal plane, and $\mathbf{K}_w$ is the PD gain matrix.

\subsection{HV-wLIP Balancing}
While the conditions given in Section \ref{sec:wipconsis} guarantees the consistence of our WBR with HV-wLIP, its balance should be maintained by tracking the desired $\delta\ddot{ x}^d$, which could be obtained by substituting the $\tau_w$ into \eqref{eq:clsys} as
\begin{equation}
	\delta\ddot{ x}^d=[0,1,0](\mathbf{A}(z)\mathbf{e}_{\text{fb}}+{\mathbf{B}(z)}\tau_w),
\end{equation}
where $\mathbf{e}_{\text{fb}}$ is the feedback value. The tracking task is implemented with following task formulation
\begin{equation}
	\mathbf{J}_\text{HV-wLIP}\ddot{\mathbf{q}}=\delta\ddot{x}^d - \dot{\mathbf{J}}_\text{HV-wLIP}\dot{\mathbf{q}}.
	\label{wlip_balance}
\end{equation}
where $\mathbf{J}_\text{HV-wLIP}$ is the Jacobian of $\dot{\delta x}$.

\subsection{Weighted WBC}
We use WBC to bridge the gap between the reduced-order model and the full dynamics of the WBR. Instead of HQP \cite{Bellicoso2016HQP, tmechhqp} or hard-constrained QP which are sensitive to task parameters and might cause feasibility issues, we chose to implement the WBC as a weighted QP inspired by \cite{Yanran_mit_humanoid}. All tasks in equality form as $\mathbf{J}\ddot{\mathbf{q}}=\mathbf{b}$ are written as a weighted sum of square
\begin{subequations} \label{weighted_QP}
\begin{align}
    [\ddot{\mathbf{q}^*},{\boldsymbol{\tau}}^*]=&\argmin_{
        \ddot{\mathbf{q}},{\boldsymbol{\tau}},{}_C\mathbf{F}_c,
        s,\tau_{w}}
    w_ss^2+\sum_{i=0}^{N_t-1}||\mathbf{J}_i\ddot{\mathbf{q}}-\mathbf{b}_i||_{\mathbf{W}_i}^2\label{eq:qp}\\
    \text{s.t.}\quad&\text{EoM in (\ref{eq:EoM})},\\
    &\boldsymbol{\tau}\in[\boldsymbol{\tau}_{\min},\boldsymbol{\tau}_{\max}],\label{eq:tqbd}\\
    &\text{Linearized friction cone in }  \eqref{eq:friccon}\label{eq:fric},\\
    &\text{Relaxed CLF constraint in } \eqref{eq:relaxedclfineq}\label{eq:relaxedclf},
\end{align}
\end{subequations}
where $(\cdot)^*$ is the optimized variable, $N_t=5$ is the number of tasks, $\mathbf{J}_i$ is the $i$th task Jacobian, $\mathbf{b}_i$ is the desired task space acceleration including bias, $\mathbf{W}_i$ is the task weight. Constraints \eqref{eq:EoM} is the full dynamics,  (\ref{eq:tqbd}) is the box constraints for the driven torques, \eqref{eq:fric} is the linear approximation of the friction cone of each contact point defined in \eqref{eq:friccon}, while \eqref{eq:relaxedclf} denotes the relaxed CLF condition defined in \eqref{eq:relaxedclfineq}.

The final commanded torques to the joints are
\begin{equation}
    \boldsymbol{\tau}_\text{cmd}=\boldsymbol{\tau}_j^*+\textbf{K}_{j,d}[\text{CLIP}(\small\sum\ddot{\mathbf{q}}_j^*\Delta t)-\dot{\mathbf{q}}_j],
\end{equation}
where $\boldsymbol{\tau}_j^*,\ddot{\mathbf{q}}_j^*$ are respectively the joint parts of $\tau^*,\ddot{\mathbf{q}}^*$, $\textbf{K}_{j,d}$ is the joint damping gain of the embedded controller at 10kHz which can enhance robustness against modeling error for sim2real and $\text{CLIP}$ denotes a clip function that limits the accumulated velocity error to prevent overshooting.

\section{Experimental Evaluation} \label{sec:experiment}
In this section, we evaluate the contributions proposed in this paper by simulation and hardware experiments.

\subsection{Hardware Implementation}
\begin{table}[t]
	\centering
	\caption{Hardware Parameters}
	\label{tab2}
	\setlength{\tabcolsep}{2.5mm}{\begin{tabular}{llll}
			\hline\hline\noalign{\smallskip}	
			\textbf{Item} & \textbf{Type} & \textbf{Rated Spec} & \textbf{Volt} \\
			\noalign{\smallskip}\hline\hline\noalign{\smallskip}
			Low-level controller & STM32F407 & /\ & 24V\\
			High-level PC & NUC11PAQi7 & /\ & 19V\\
			IMU & BMI-088 & /\ &\\
			Joint motors & 8115-1:6 & 13-35Nm,120rpm & 48V\\
			Wheel motors & DM6006 & 3.4-12.5Nm,300rpm& 48V\\
			
			\noalign{\smallskip}\hline\hline
	\end{tabular}}  
	\label{tableii}
\end{table}

An overview figure of the constructed hardware system is shown in Fig. \ref{fig:overview_first}. The main structure of the robot consists of carbon fiber sheets, 3D printed parts and aluminum alloy connectors. 
The robot is powered by a 12S2P 48V 8000mAh 21700 Li-ion array connected to two voltage regulators to output 48V and 24V for different devices. The motors communicate with a low-level controller via CAN2.0b@1Mbps while the low-level controller communicates with an Intel NUC via USB2.0FS@ 500Hz. Hardware details are shown in Table. \ref{tableii}.

\subsection{Velocity Regulation Performance of HV-wLIP} \label{sec:model}

To evaluate the performance advantage of HV-wLIP in terms of velocity regulation, we deployed the WBC \eqref{weighted_QP} in numerical simulation to execute post-impact balancing. The WBR started from a fixed initial joint configuration $\mathbf{q}_j^0$ and base pose at a non-zero base tangential velocity, and the controller was commanded to stop moving and restore balance. We use RaiSim \cite{raisim} for simulation, and the weights in \eqref{eq:ocp} are $\mathbf{Q}=\text{diag}[10, 1, 1], R=0.01.$
Other relevant parameters of WBC are given in Table \ref{tablesim}. For comparison, we employ a WIP-based WBC similar to \eqref{weighted_QP} for the same experiment but with different tasks including an LQR task $\tau_w=-\mathbf{K}_\text{WIP}[\dot{x}_c,\dot{\vartheta},\vartheta]^\top$ and a WIP constraint task mapped to the joint space to enforce fixed body by $\ddot{\mathbf{q}}_j=\mathbf{K}_\text{j,wip}[(\mathbf{q}_j^0-\mathbf{q}_j)^\top, -\dot{\mathbf{q}}_j^\top]^\top,$ where the $\mathbf{K}_\text{WIP}$ is the LQR gain computed with the linearization of \eqref{eq:wip} and $\mathbf{K}_\text{j,wip}$ is PD gain matrix.
\begin{table}[t]
	\centering
	\caption{Control Parameters}
	\setlength{\tabcolsep}{2.5mm}{\begin{tabular}{lllll}
			\hline\hline\noalign{\smallskip}	
			\textbf{Task} & $K_p$ & $K_d$ & $W$ \\
			\noalign{\smallskip}\hline\hline\noalign{\smallskip}
			CoM height $z$ \eqref{eq:comz_task} & 100 & 10 & 100 \\
			Wheel departure \eqref{eq:sync} & 1000 & 30 & 10 \\
			Base pitch \eqref{pel_pit} & 100 & 10 & 1\\
			Base yaw  \& roll \eqref{yaw_roll} & [100, 100] & [10, 10] & diag([10, 10])\\
			CLF slackness $w_s$ \eqref{eq:qp} & / & / & 1000 \\
			HV-wLIP balancing \eqref{wlip_balance} & / & / & 10 \\
			\noalign{\smallskip}\hline\hline
	\end{tabular}}  
	\label{tablesim}
\end{table}

\begin{figure}[t!]
	\centering
	\includegraphics[width=\linewidth]{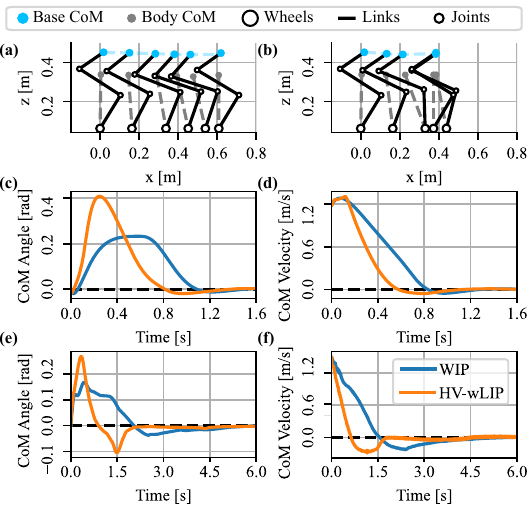}
	\caption{Simulated trajectory of robots controlled WIP and wLIP model. (a) Result of WIP-based WBC. (b) Result of the wLIP-based WBC. (c-d) Trajectories of the CoM tilting angle and the velocity in simulation. (e-f) Trajectories of the CoM tilting angle and the velocity in hardware experiment.
	}
	\label{fig:wipvswlip}
\end{figure}

The simulated results are shown in Fig. \ref{fig:wipvswlip}(a,b). It is clearly shown that robot controlled with HV-wLIP could stop within a shorter distance of 0.38m, while the WIP model provided longer stopping distance at 0.60m. As shown in Fig. \ref{fig:wipvswlip}(c,d), the HV-wLIP has smaller excessive angular momentum and thus allows faster motion, i.e., larger tilting angle $\vartheta$. In contrast, the WIP-controlled robot suffers from the coupled body and the linearization error and responds slower.

Similar task has been tested in the hardware system; instead of the initial speed, we try to kick the WBR system to make it starting from even higher CoM velocity $1.5$m/s. The experimental results in Fig. \ref{fig:wipvswlip}(e,f) show identical conclusion as in simulation, please refer to the attached video for a clear demonstration.

\subsection{Squatting of WBR} \label{sec:sim_squat}
As is mentioned in Section \ref{sec:bio}, inspired by human bio-mechanics, we aim to balance the torque distribution in hip and knee joints during squatting task. 
\subsubsection*{Inverse Kinematics}
We use the following inertia-aware inverse kinematics (IK) to generate a mapping between joint configurations and $z^d$ satisfying the torque ratio:
\begin{equation}\label{eq:ik}
		\boldsymbol{\Theta}^*(z^d)=\underset{\boldsymbol{\Theta}\in\mathcal{O}}{\argmin}\;\;|r_\tau-1|,~\text{s.t.}~\eqref{eq:effectivenessproof},
\end{equation}
which is solved by IPOPT \cite{ipopt} for a range of $z^d$ and fitted by piece-wise polynomials.

\begin{figure*}[t]
	\centering
	\includegraphics[width=0.80\textwidth]{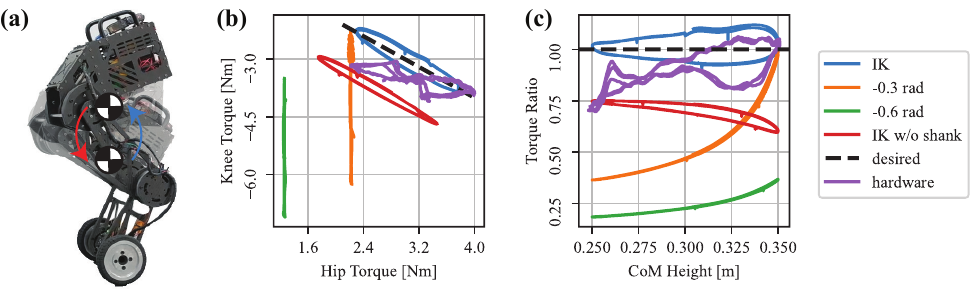}
	\caption{Squat experiment. (a) Squat motion illustration. (b) Hip and knee torque relation. (c) Torque ratio $r_\tau$ during squatting. The green and yellow lines denote constant $\theta_P^d$ tracking control at $-0.6$, $-0.3$rad respectively. The blue line denotes the proposed control strategy \eqref{eq:ik} and the red line denotes \eqref{eq:ik} without shank inertia. The purple line denotes the hardware experiment result. The black dashed line denotes the desired curve where $r_\tau=1$.}
\label{fig:sim_tq}
\end{figure*}
\subsubsection*{Comparison with alternatives}
We control the proposed WBR with the proposed controller \eqref{weighted_QP} and compare with other alternatives in simulation. Previous works commonly track constant $\boldsymbol{\Theta}$ \cite{bitwiptmech}. In our test, $\theta_P=-0.3$, $-0.6$rad are selected.
To verify the importance of precise inertia information, we neglect the shank inertia to form an inaccurate IK and compute the resultant $r_\tau$.
For all examples, the WBR is commanded to track $z^d(t)=0.3+0.05\sin(t)\in[0.25,0.35]$m. The simulation results are summarized in Fig. \ref{fig:sim_tq}.

We first note that the constant pitch angle tracking strategies, $-0.3$rad (yellow line) and $-0.6$rad (green line) show poor performance in balancing torque distribution: the hip torque is nearly constant in Fig. \ref{fig:sim_tq}a as the hip moment arm is always the same, while the torque ratios are much lower than 1, indicating that the knee motor outputs larger torque than the hip. 
The inaccurate IK shows improvement over constant $\theta_P$ by an approximate $r_\tau\in [0.6,0.75]$, but is still far from the ideal case because the imprecise inertia resulted to a steady error in torques, i.e., an $\approx1.2$Nm knee torque shift away from the desired unitary ratio (black dashed line). In contrast, the proposed strategy (blue line) could enable a steady trajectory around the ideal ratio, indicating that our IK can provide accurate kinematics reference and our bionic design can achieve human-like, balanced torque distribution.

\subsubsection*{Experimental validation}
To further demonstrate the validity of our design and control strategy, the designed WBR robot is commanded to perform the same squat task. The experimental results are summarized in Fig. \ref{fig:sim_tq} as purple line. Without mechanical parameter identification, due to the inevitable inertia mismatch, an error is induced between experimental and simulation results. Nevertheless, the experiment result exhibits improvement compared to all other alternatives by being closer to the desired ratio (black dashed line).
The experiment was recorded in the attached video for reference.

\subsection{Dynamic Performance Test} \label{sec:tracking_test}
\begin{figure*}[t]
	\centering
	\includegraphics[width=0.8\textwidth]{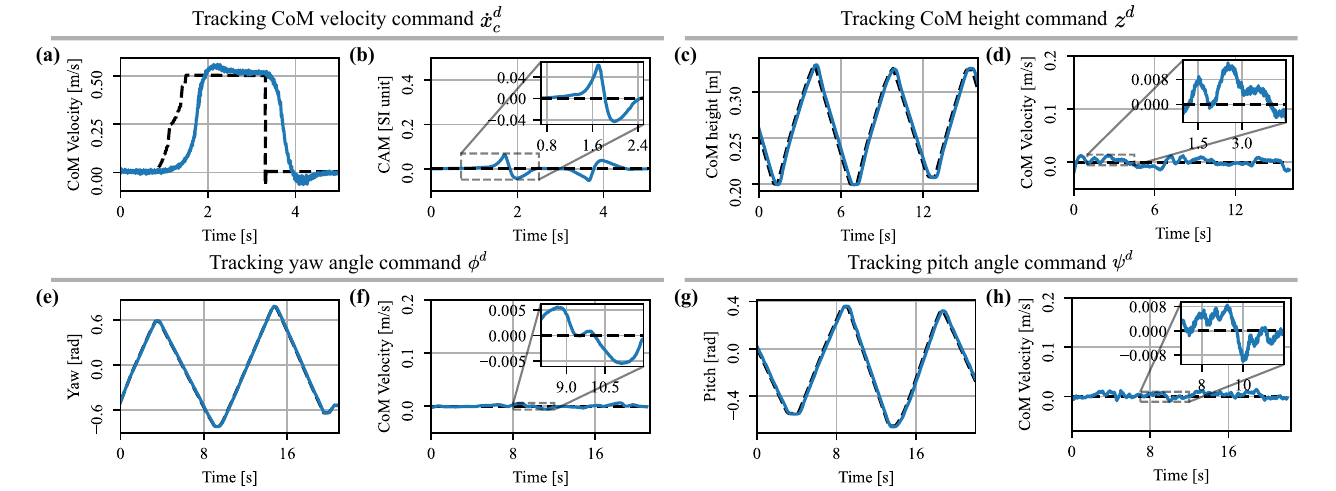}
	\caption{Dynamic performance test. (a-b) The CoM velocity and CAM data in tracking $\dot{x}^d_c$. (c-d) The CoM height and velocity data in tracking $z^d$. (e-f) The yaw and CoM velocity data in tracking $\phi^d$. (g-h) The pitch and CoM velocity data in tracking $\psi^d$. The dashed lines are user commands.}
	\label{fig:base}
\end{figure*}

\begin{figure*}[t]
	\centering
	\centering
	\includegraphics[width=\textwidth]{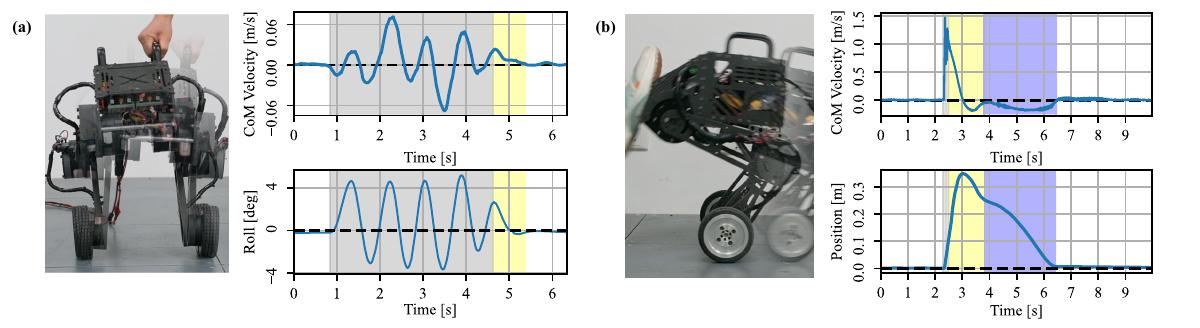}
	\caption{Robustness against unknown external disturbance. (a) Lateral shaking and the recorded data of CoM velocity $\dot{x}_c$ and roll angle $\psi$. (b) Kick impact and the recorded data of CoM velocity $\dot{x}_c$ and position. The dashed black lines denote desired states. The gray area marks the moment when external disturbance is imposed, the yellow area marks the balancing while the cyan span marks the resuming to original position. 
	}\label{fig:fig_anti}
\end{figure*}

We demonstrate the dynamic performance of the proposed tracking controller \eqref{weighted_QP} on tracking user-commanded CoM velocity $\dot{x}_c^d$, CoM height $z^d$, base orientation yaw $\phi\in\mathbb{R}$, pitch $\theta\in\mathbb{R}$. The experiment results are shown in Fig. \ref{fig:base} and recorded in the attached video material. Due to the limitation of mechanical structure, active tracking of varying roll $\psi$ cannot be implemented. Instead, its passive regulation in balance is tested in next section. 

It is shown that the controller could effectively track the user commands. For the velocity task, the CAM was well-regulated within 0.04, which confirms the model consistency presumption in Section \ref{sec:wipconsis}. The other pose tasks show good tracking performance while maintaining the balance with a small $\dot{x}_c$ variation within 0.01m/s.

\subsection{Robustness against External Disturbance}

\subsubsection*{Random external disturbance}
To test the robustness of controller \eqref{weighted_QP} against external disturbance, we commanded the robot to stand still, and then 1) impose manual lateral shaking on the robot to test the compliance and stability of roll angle regulation, 2) kick the robot to test its stability under impact. The experiment results are shown in Fig. \ref{fig:fig_anti} and recorded in the attached video for demonstration.

For the lateral shaking shown in Fig. \ref{fig:fig_anti}a, the roll angle was perturbed around $10^\circ$. After released, the robot resumed zero velocity with less than $0.8$s and the WBR was well-balanced during the whole procedure. For the impact test shown in Fig. \ref{fig:fig_anti}b, the WBR was kicked heavily. The controller spent around $1.5$s to stop and returned the original position after $2.2$s with a maximum displacement of $0.33$m. The robustness against random external disturbance is clearly demonstrated.

\subsubsection*{Unknown terrain}

\begin{figure*}[t]
	\centering
	\includegraphics{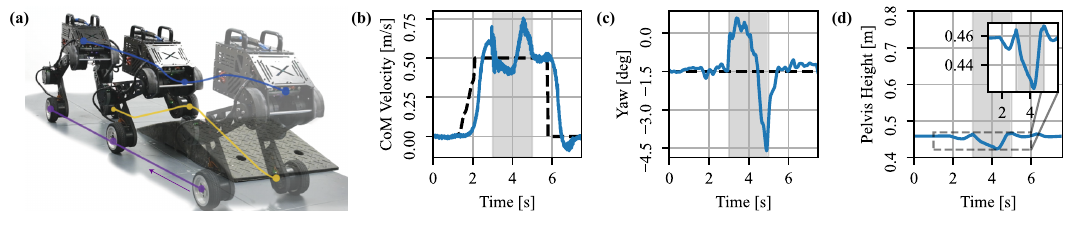}
	\caption{Robustness against unknown terrain. (a) The demonstration of the experiment where the WBR was commanded to pass a pair of highly inclined, symmetrically placed slopes. 
	(b-d) The measured data of CoM velocity $\dot{x}_c$, yaw $\phi$, and base height when passing the slopes. Duration of passing the slopes is marked with gray span in each figure where the black dashed lines denote the tracking commands. 
	}
	\label{fig:clmb}
\end{figure*}

To test the compliance and robustness against unknown terrain, the WBR was controlled to pass a pair of highly inclined symmetric slopes. During the process, the desired CoM velocity was set $0.5$m/s and a constant pose was maintained. The experiment procedure is demonstrated in Fig. \ref{fig:clmb}a and the measured data is shown in Fig. \ref{fig:clmb}b-d, with recorded clip included in the attached video material. 

It is shown that upon in contact with the slope, a jump in the CoM velocity and pose happened while the controller attempted to regulate them. Due to the un-modeled non-vertical GRF, the tracking of velocity and yaw were not perfect but were still acceptable. Specifically, the velocity was controlled around the desired value and the error of yaw are no bigger than $5^\circ$. After passing the slopes, they were quickly well tracked again.
Other pose tasks, including roll, pitch and CoM height, were well maintained. Please refer to the attached video for clear demonstration.
Furthermore, we observed that when climbing over the slope, the base height, shown in Fig. \ref{fig:clmb}d, was actively lowered such that the CoM height could be kept constant, such phenomenon reveals the effectiveness of the proposed controller \eqref{weighted_QP}.

\begin{remark} [Computational efficiency]
	The WBC \eqref{weighted_QP} is implemented using CasADi \cite{casadi} and OSQP \cite{osqp}. The total average time for computing the WBC, including solving the infinite-horizon OC \eqref{eq:ocp}, is measured around $220\mu$s. As the control frequency of hardware is set $500$Hz and the corresponding sampling interval $T_s$ is $2$ms, the computation is only $\approx 10$\% of $T_s$, which is efficient for real-time control of the proposed WBR system.
\end{remark}

\section{Conclusion}\label{sec:conclu}
In this paper, a serial-legged wheeled bipedal robot is designed based on bio-mechanic analysis of human squatting, aiming to improve the torque effectiveness of hip and knee motors without losing the flexible base workspace. It is controlled using a CLF-based WBC with HV-wLIP. As a better approximation of WBR locomotion by considering non-fixed body, variable height and wheel torque, the proposed model is demonstrated to have superior dynamic performance than WIP. The proposed CLF condition provides theoretically guaranteed stability in the presence of various constraints and task-space objectives. Hardware experiments are presented to demonstrate the advantages of the proposed hardware and controller design. For future work, the robot can be loaded with robotic arms or torso to further extend the capability and utilize the advantage of our bionic design.

\bibliographystyle{IEEEtran}
\bibliography{IEEEabrv,reference}
\vskip -2\baselineskip plus -1fil
\begin{IEEEbiography}[{\includegraphics[width=1in,height=1.25in,clip,keepaspectratio]{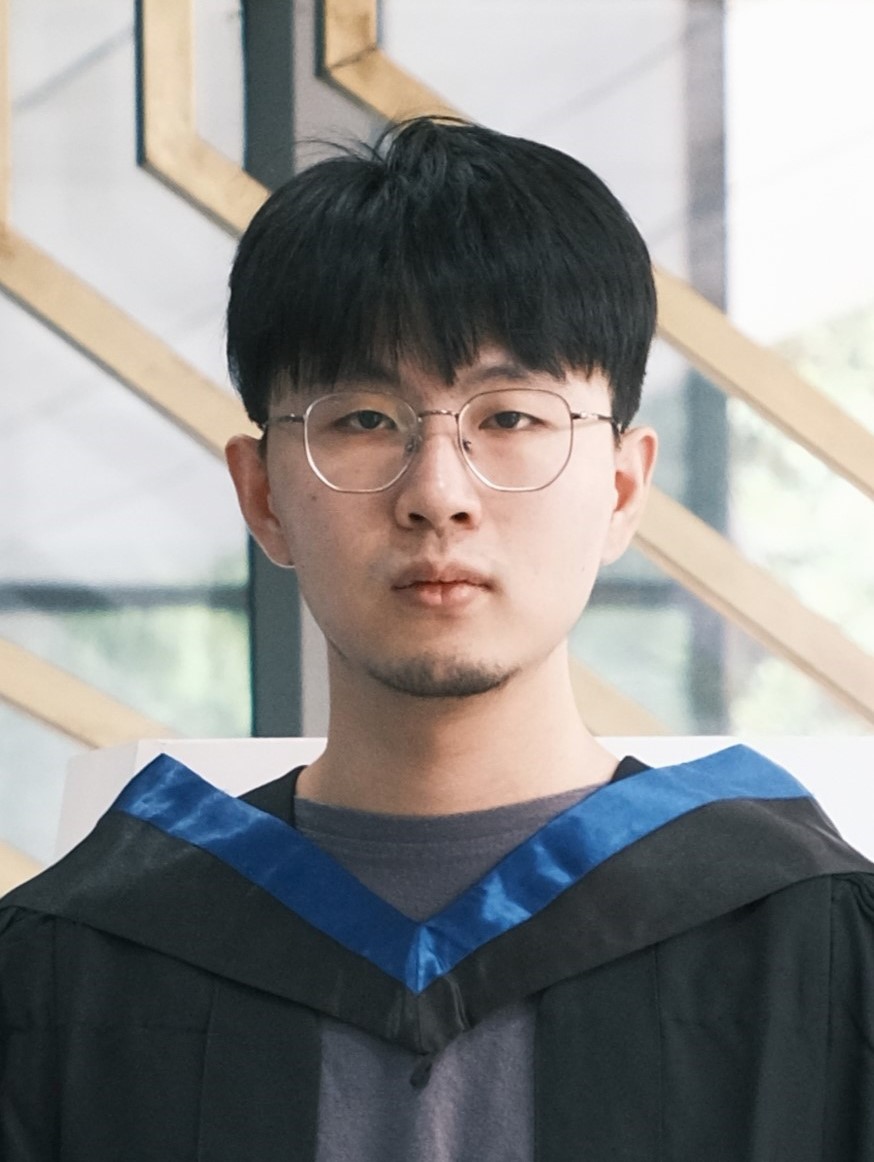}}]{\textbf{Haizhou Zhao}} received the B.Eng. degree in Mechatronics and Robotic Systems from  Xi'an Jiaotong-Liverpool University, Suzhou, China, in 2023. His research is focused on robotic control and mechanical design.
\end{IEEEbiography}
\vskip -2\baselineskip plus -1fil
\begin{IEEEbiography}[{\includegraphics[width=1in,height=1.25in,clip,keepaspectratio]{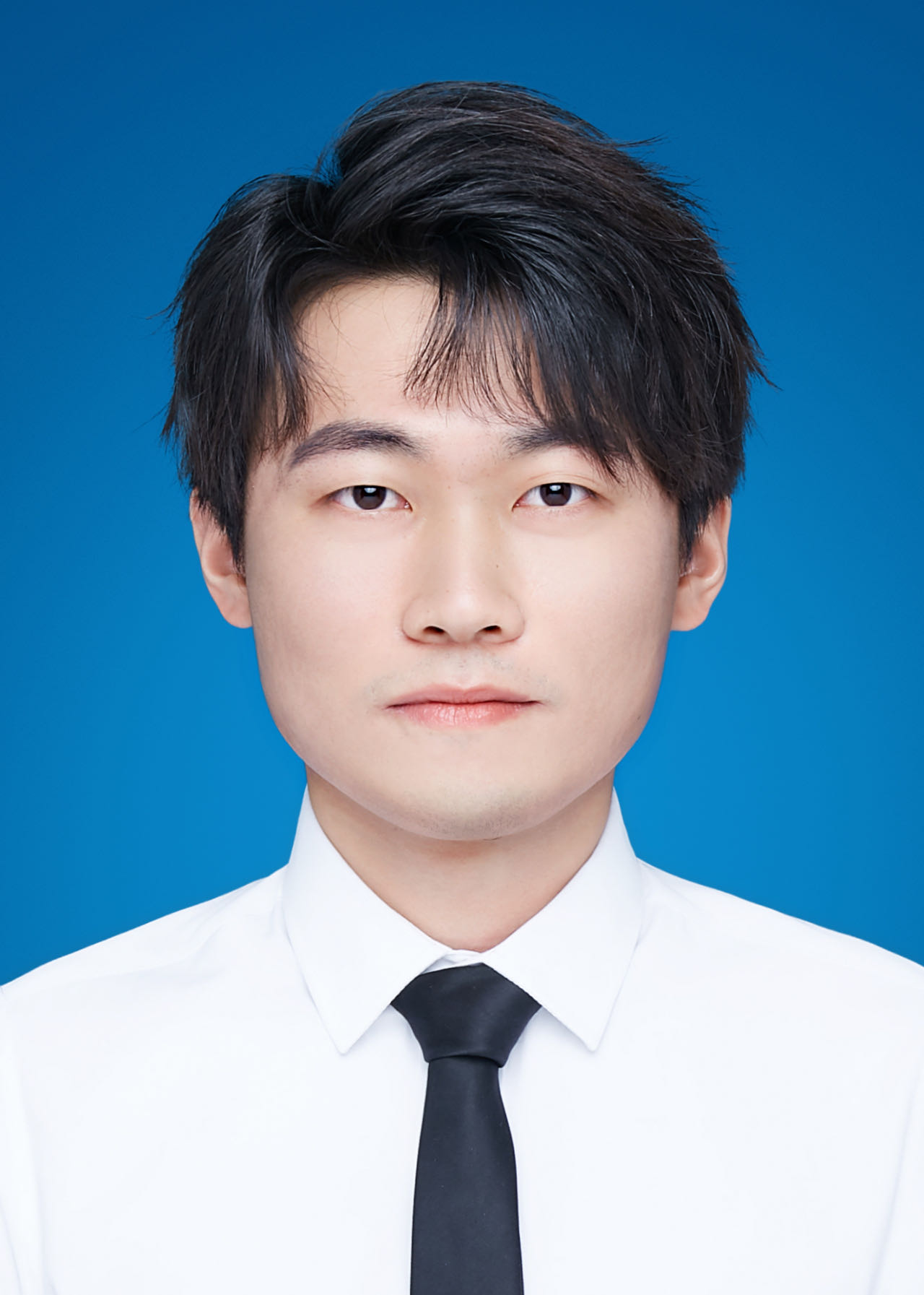}}]{\textbf{Lei Yu}} received the B.Eng. degree in Vehicle Engineering, in 2020, and the M.Res. degree in Pattern Recognition and Intelligent Systems from the University of Liverpool, Liverpool, UK, in 2022. He is currently working toward the Ph.D. degree in Electronic and Electrical Engineering with the University of Liverpool.	His research interests include variable stiffness actuator, bionic robots, and optimal control.
\end{IEEEbiography}
\vskip -2\baselineskip plus -1fil
\begin{IEEEbiography}[{\includegraphics[width=1in,height=1.25in,clip,keepaspectratio]{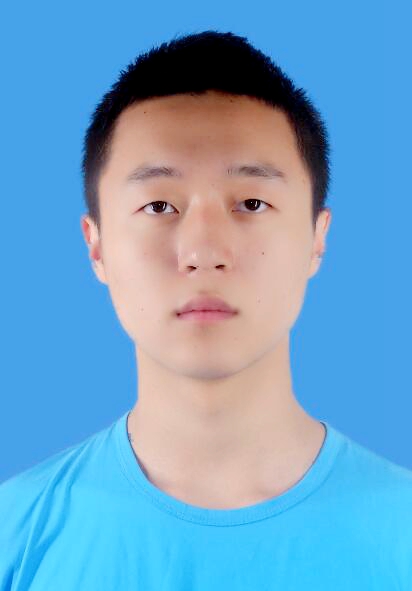}}]{\textbf{Siying Qin}} received the B.Eng. degree in Mechatronics and Robotic Systems from the Xi'an Jiaotong-Liverpool University, Suzhou,China, in 2022. He is currently working toward the Ph.D. degree in Electronic and Electrical Engineering with the University of Liverpool, Liverpool, UK. His research is focused on optimal control and legged robot.
\end{IEEEbiography}
\vskip -2\baselineskip plus -1fil
\begin{IEEEbiography}[{\includegraphics[width=1in,height=1.25in,clip,keepaspectratio]{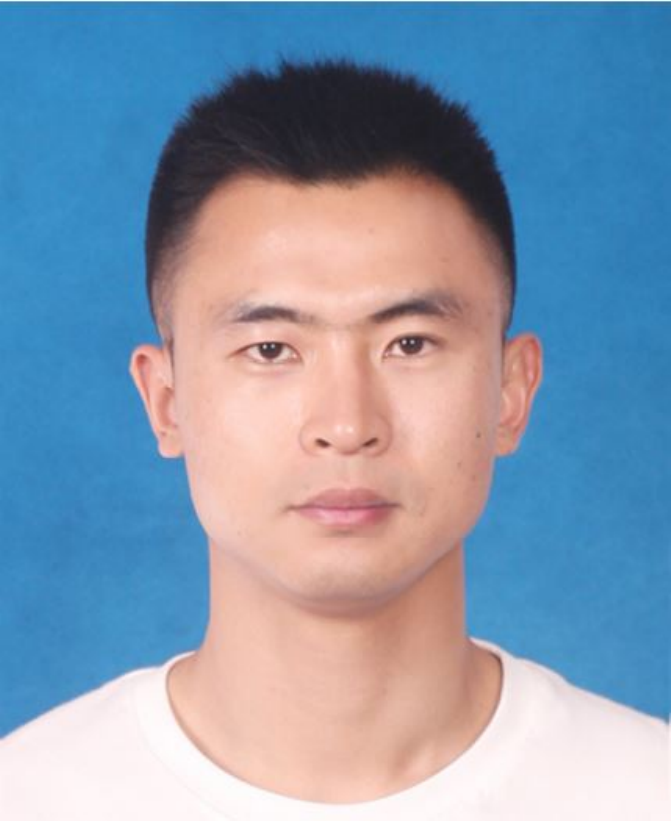}}]{Gumin Jin}
	received the B.S. degree from Harbin Institute of Technology, Harbin, China, in 2013. He is currently pursuing the Ph.D. degree with the Department of Automation, School of Electronic Information and Electric Engineering, Shanghai Jiao Tong University , China. His research interests include computer vision, 3-D sensing and metrology, and robotics.
\end{IEEEbiography}
\vskip -2\baselineskip plus -1fil
\begin{IEEEbiography}[{\includegraphics[width=1in,height=1.25in,clip,keepaspectratio]{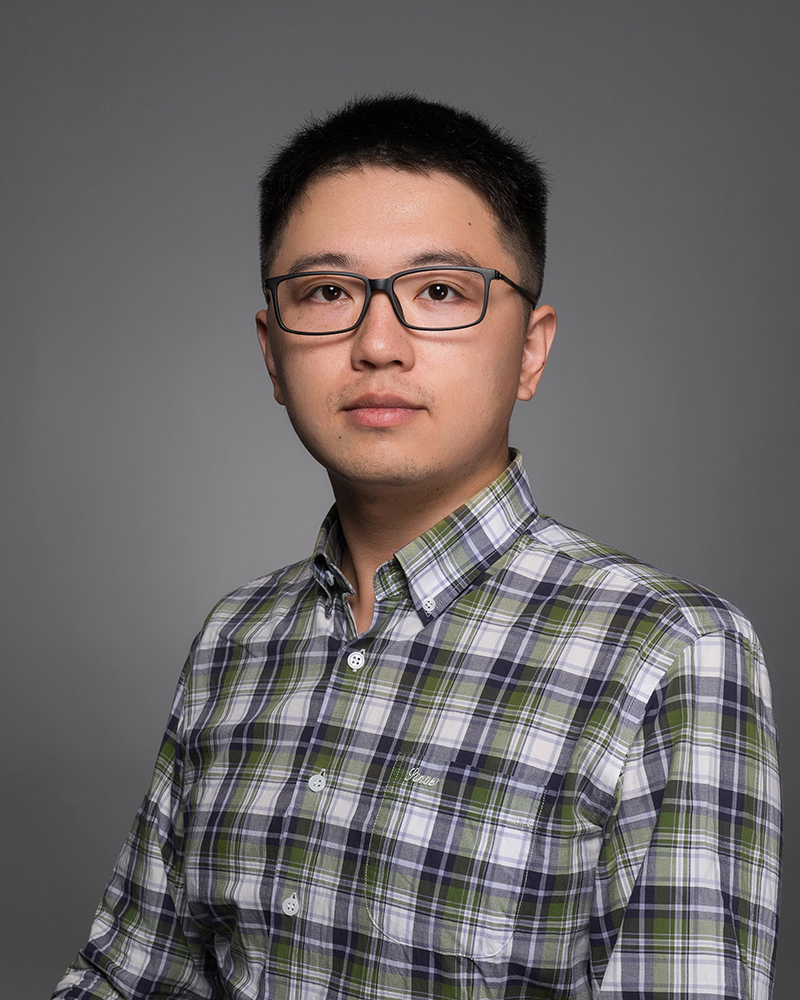}}]{\textbf{Yuqing Chen}} received his Ph.D. degree from Singapore University of Technology and Design (SUTD) in 2020, M.Eng. and B.Eng. degrees in Control Science and Engineering from Harbin Institute of Technology (HIT) in 2015 and 2013 respectively. He is currently an Assistant Professor with the Department of Mechatronics and Robotics, School of Advanced Technology, Xi'an Jiaotong-Liverpool University. His main research interests include robot control, optimal control of dynamical systems and hardware-in-the-loop optimal control theory.
\end{IEEEbiography}

\end{document}